\pdfoutput=1

\documentclass[11pt]{article}
\usepackage{authblk}
\usepackage{acl}
\usepackage{xspace}
\def\datasetname{CF-TriviaQA}
\def\factualdatasetname{F-TriviaQA}
\usepackage{times}
\usepackage{latexsym}
\usepackage{amsmath}
\usepackage{makecell}

\usepackage[T1]{fontenc}

\usepackage[utf8]{inputenc}
\usepackage{booktabs}
\usepackage{microtype}
\usepackage{graphicx}

\title{Hallucination Augmented Recitations for Language Models}

\author{Abdullatif Köksal${}^{1, 2}$\thanks{\protect\space\space Work performed during an internship at Google.} \quad Renat Aksitov${}^3$ \quad Chung-Ching Chang${}^3$\\
${}^1$Center for Information and Language Processing, LMU Munich \\
\ ${}^2$Munich Center of Machine Learning \ ${}^3$Google Research \\
\texttt{akoksal@cis.lmu.de} \texttt{\{raksitov, ccchang\}@google.com}
}

\begin{document}
\maketitle
\begin{abstract}
Attribution is a key concept in large language models (LLMs) as it enables control over information sources and enhances the factuality of LLMs. While existing approaches utilize open book question answering to improve attribution, factual datasets may reward language models to recall facts that they already know from their pretraining data, not attribution.
In contrast, counterfactual open book QA datasets would further improve attribution because the answer could only be grounded in the given text.
We propose \textbf{Hallucination Augmented Recitations (HAR)} for creating counterfactual datasets by utilizing hallucination in LLMs to improve attribution.
For open book QA as a case study, we demonstrate that models finetuned with our counterfactual datasets improve text grounding, leading to better open book QA performance, with up to an 8.0\% increase in F1 score.
Our counterfactual dataset leads to significantly better performance than using human-annotated factual datasets, even with 4x smaller datasets and 4x smaller models. We observe that improvements are consistent across various model sizes and datasets, including multi-hop, biomedical, and adversarial QA datasets.
\end{abstract}

\section{Introduction}

Text grounding or attribution\footnote{We use these terms interchangeably throughout the paper.} is a key aspect in large language models (LLMs). Since most LLMs are trained once without any update, attributability gives LLMs an adaptation ability to dynamic changes in the real world, such as temporal questions \cite{vu2023freshllms}. Additionally, attribution improves the factuality of language models and could help to control the source of information more granularly \cite{gao-etal-2023-rarr}. Recent works in LLMs focus on adapting retrieval-augmented approaches, such as search-engine-assisted systems like Bard and GPT-4, to utilize attributable LLMs \cite{NEURIPS2020_6b493230, chen-etal-2017-reading}.

\begin{figure}[t!]
    \centering
    \includegraphics[width=\linewidth]{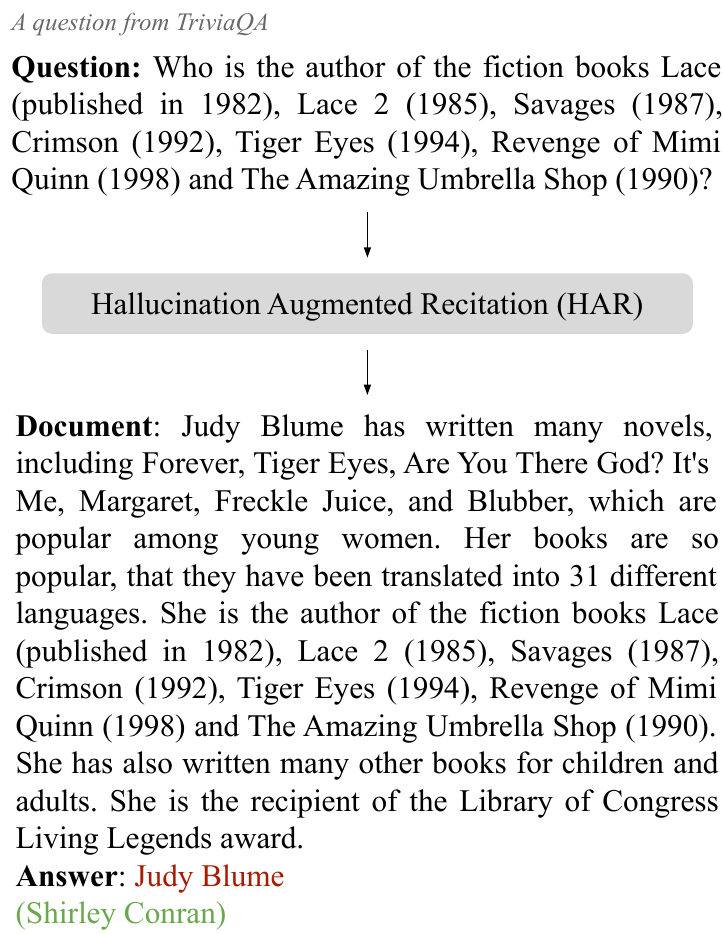}
    \caption{A counterfactual example from \datasetname, generated via the HAR pipeline. For a given question from TriviaQA, we utilize LLM hallucination to generate high-quality, attributable, and counterfactual open book QA examples. As illustrated in this example, our HAR pipeline outputs a hallucinated\protect\footnotemark\space document supporting the counterfactual answer of Judy Blume, while the gold answer is Shirley Conran as given in TriviaQA.
    }
    \label{fig:har_main}
\end{figure}
\footnotetext{LLM's hallucination in this example is likely due to the ambiguous book name ``Tiger Eyes'', as both Judy Blume and Shirley Conran have authored books with that title. However, Judy Blume's Tiger Eyes was published in 1981, and she did not write any books with the other names listed in the question.}

To improve the text grounding abilities of language models, most works focus on including open book question answer (QA) tasks which include attributable documents for given questions. Therefore, the largest tasks in the training datasets of recent instruction-tuned models such as T\textit{k}-Instruct \cite{wang-etal-2022-super} or InstructGPT \cite{NEURIPS2022_b1efde53} are based on open book QA.  However, there is an underlying multi-objective trade-off in open book QA. Since pretrained language models already know facts through their pretraining data, finetuning them with factual open book QA datasets could reward the model to recall the facts from their memory without attribution instead of attributing to the document. Therefore, attribution and recall are competing factors in language models because facts in open book QA datasets may already present in the memory of language models. In contrast, counterfactual data mitigate such spurious correlations and further improve the text grounding abilities of language models \cite{Kaushik2020Learning}. This is because counterfactual data introduces a conflict with the memory, preventing straightforward recall. However, recent work focusing on counterfactual open book QA via entity substitution or retrieval-based generation demonstrates only limited and inconsistent improvements in text grounding \cite{longpre-etal-2021-entity, paranjape-etal-2022-retrieval}.

We propose \textbf{Hallucination Augmented Recitations (HAR)} which utilizes LLM hallucination\footnote{We define hallucination as the generation of counterfactual text conflicting with real-world knowledge.} to create a counterfactual open book QA dataset. HAR builds on the recitation augmentation \cite{sun2023recitationaugmented} by prompting LLMs to introduce reasoning through recitation and produce an attributable document for a given question. This produces high-quality and consistent counterfactuals, in contrast to heuristics such as entity replacement \cite{longpre-etal-2021-entity}. We then apply additional filters to identify high-quality hallucinations from LLMs and create a counterfactual dataset. In Figure \ref{fig:har_main}, we illustrate an example from our dataset, \datasetname. For a given question about an author of several books, the LLM hallucinates and generates a counterfactual document that supports an incorrect answer (Judy Blume). Therefore, a model finetuned on such counterfactual open book QA dataset can be rewarded only by attributing to the document, since the answer cannot be recollected from the model's memory. %

Our contributions are as follows:
\begin{enumerate}
    \item We utilize hallucination and propose the \textbf{Hallucination Augmented Recitations (HAR)} pipeline to create a high-quality attributable counterfactual open book QA, named \textbf{\datasetname{}} with 19K examples.
    \item We show that T5 models finetuned with \datasetname{} significantly outperform those finetuned with human-annotated factual open book QA datasets, even \textbf{with 4x smaller datasets and 4x smaller model sizes}.
    \item We observe that our findings are consistent across various model sizes, ranging \textbf{from 60M to 11B}, and on various datasets, including \textbf{multi-hop}, \textbf{biomedical}, or \textbf{adversarial} questions. 
\end{enumerate}

\begin{figure*}[t]
    \centering
    \includegraphics[width=\linewidth]{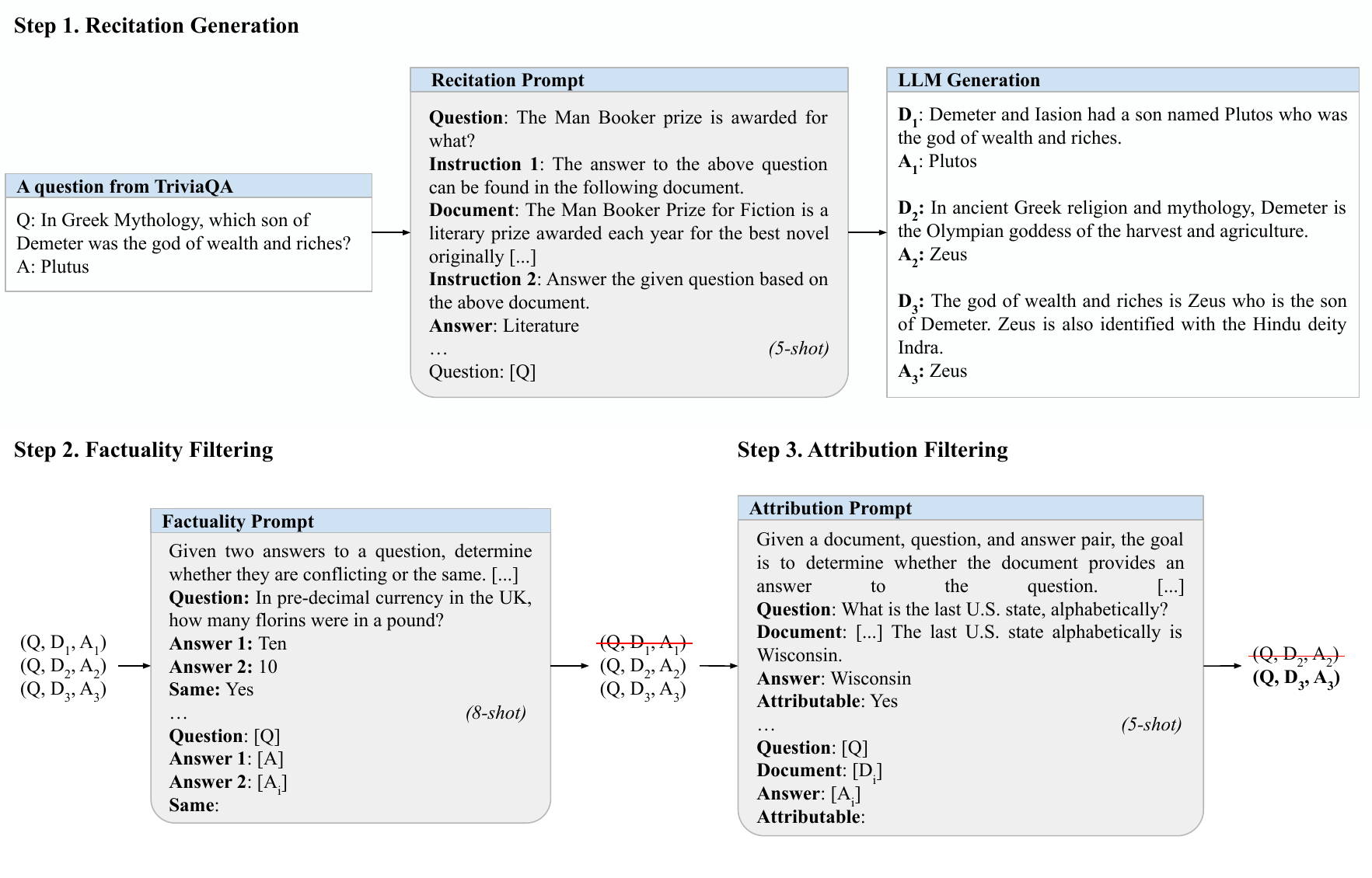}
    \caption{Three steps of HAR. HAR first generates document, and answer pairs for a given question. Then, it filters factual generations (e.g., the answer with Plutos while the gold answer is Plutus). Finally, it filters generated examples without text grounding (e.g., the second document-answer pair, where there is no mention of Zeus in the document). See full prompts in \S\ref{sec:appendix_prompt}.}
    \label{fig:har_pipeline}
\end{figure*}

\section{Hallucination Augmented Recitations}
We aim to create a counterfactual question answering dataset to further improve
attribution in language models. To this end, we utilize hallucination for
counterfactuals and propose Hallucination Augmented Recitations (HAR). HAR has
three steps as illustrated in Figure \ref{fig:har_pipeline}.
\begin{enumerate}
    \item \textbf{Recitation generation}: We use the recitation-augmented language model approach \cite{sun2023recitationaugmented} to generate multiple document and answer pairs for a given question. 
    \item \textbf{Factuality Filtering}: We filter out factual answers to focus on counterfactuality.
    \item \textbf{Attribution Filtering}: We apply the attribution filter to remove question, document, and answer pairs where the answer is not grounded in the document.
\end{enumerate}

\subsection{Recitation Generation}
Recitation generation is the first step of HAR, as described in Figure \ref{fig:har_pipeline}. We follow \citet{sun2023recitationaugmented} and apply the
recitation-augmented language model approach. Simply, we generate multiple
document and answer pairs for a given question by using a 5-shot prompt via LLMs
to create open book QA examples. We design the recitation prompt as shown in Figure
\ref{fig:recitation_prompt} in the Appendix which conditions LLMs to generate document, and answer pairs for a given
question. We manually pick five high-quality examples from TriviaQA
\cite{joshi-etal-2017-triviaqa} as our few-shot examples. 

During generation, we use questions only from TriviaQA
\cite{joshi-etal-2017-triviaqa}, and utilize PaLM 2-L \cite{anil2023palm} as our
LLM in recitation generation. For each question, we perform 24 iterations with
temperature sampling ($T=0.7$). Then, we parse generated documents and answers
and eliminate examples that do not follow our prompt (e.g., no new line between
document and instruction 2).

\subsection{Factuality Filtering}

Since we only focus on counterfactual examples to improve
attribution, we filter out examples that have factual answers. First, we simply remove
generated examples whose answers have the same surface form as the gold answer
in TriviaQA. However, our manual evaluation of the surface form filtering shows that there
are many generated pairs where the generated answers are factual but their surface forms are different from the gold answers. It could be caused by synonyms,
transliterations, accented characters, or reformatted answers (e.g., 1930 vs. 30s, Eugène Delacroix vs. Eugene Delacroix, or Plutos vs. Plutus as in Figure \ref{fig:har_pipeline}); therefore
heuristics that apply filtering on surface forms are not enough to filter
factual answers. In the second part of HAR, we propose a factuality
filtering method via LLMs to remove such factual answers and keep counterfactual answers
only.

We propose a PaLM 2-L based filtering method with 8-shot examples. In the prompt,
we only provide question and answer pairs without a document because we are only interested in finding whether the generated answer leads to the same answer as the gold answer for the given question. We select four examples whose generated answers are the same as the gold answers but have
different surface forms and four examples with different answers. We design
the prompt as illustrated in Figure \ref{fig:factuality_prompt} in the Appendix. We compare the probabilities
of `Yes' and `No' tokens for the next token and decide factuality based on the normalized probability of the `Yes' token.

\begin{figure*}
    \centering
    \includegraphics[width=\linewidth]{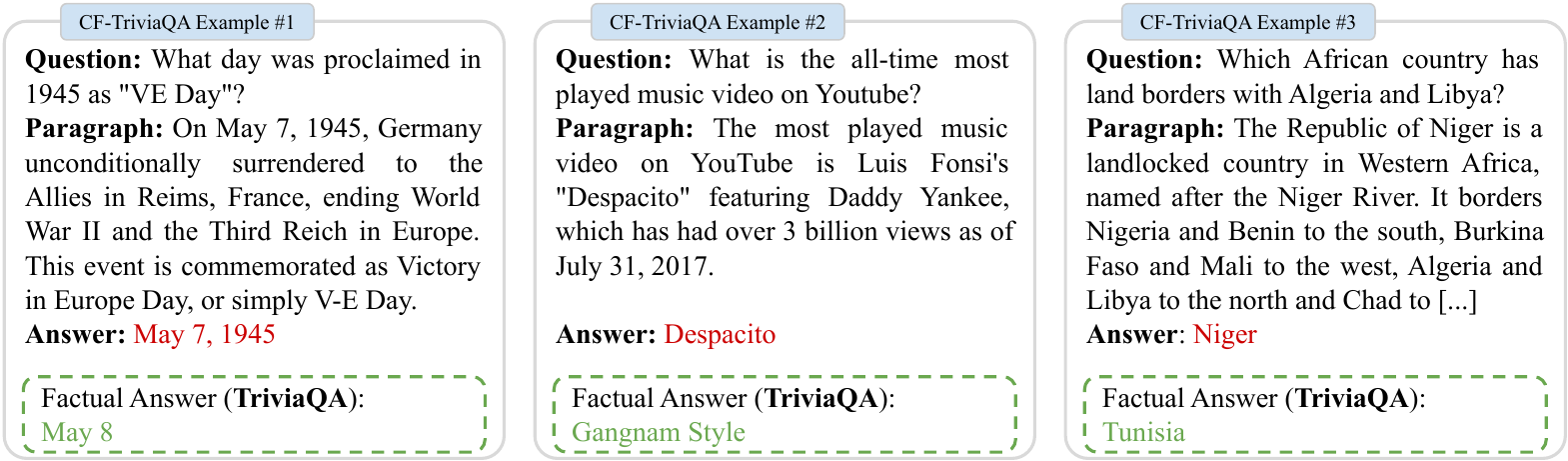}
    \caption{Three examples from \datasetname. They are counterfactual and conflict with the answers in TriviaQA, yet they have different attributes. The first example is simply counterfactual with an incorrect date, while the second example includes a temporal question in which the answer could change over time. The last example is an ambiguous question since there are two factual answers but the paragraph provides only one response.}
    \label{fig:dataset_examples}
\end{figure*}

\subsection{Attribution Filtering}

After the recitation generation and factuality filtering steps, we have a set of
question, document, and answer pairs with only counterfactual answers. During
our manual analysis of these examples, we encounter examples where the
generated answer is not grounded in the generated document. In order
to eliminate such cases, we again use filtering via LLMs (i.e., PaLM 2-L) with 5-shot
examples. We design a prompt with generated counterfactual examples with and without grounding,
as shown in Figure \ref{fig:attribution_prompt} in the Appendix.

We calculate the probability of `Yes' and `No' tokens and normalize it. If the
normalized probability of the `Yes' token is lower than $0.5$, we remove those samples
from our dataset. Furthermore, we may have different document, and answer pairs for
a given question from the HAR pipeline since we generate 24 examples for the recitation generation part. To select a unique sample for each
question, we select the document answer pair with the highest normalized attribution score.

\section{Counterfactual TriviaQA: \datasetname}

We propose \datasetname, a counterfactual dataset generated from the TriviaQA dataset \cite{joshi-etal-2017-triviaqa} using Hallucination Augmented Recitations (HAR). To construct \datasetname, we first apply recitation generation to generate 24 document, answer pairs for each question in TriviaQA, resulting in an average of approximately 3 unique answers per question. More than 30\% of the generated examples have the gold answer as a generated answer, which means they are not counterfactual. Furthermore, we observe that generated answers with different surface forms than the gold answer could be still factual (e.g., synonyms, hyponymy, translation), or the generated answer may not be grounded in the generated text.

Next, we apply the factuality filter to remove factual answers, which eliminates more than 45\% of the remaining examples. We then apply an attribution filter to remove generations without text grounding, which also removes more than 50\% of the remaining data after the factuality filtering. We also select only one context for each answer with the highest attribution score according to our filter. Finally, we obtain our counterfactual dataset, \textbf{\datasetname}, with 19,327 examples.

\datasetname{} contains different types of counterfactual examples, as illustrated in Figure \ref{fig:dataset_examples}. The first example shows a simple counterfactual example with a different date than the factual date in the gold data. We mostly see examples in this category in \datasetname, simply counterfactual. The second example illustrates the temporal aspect of counterfactuality. The gold answer was factual when TriviaQA was first published, but due to the temporal aspect of the question, the original gold answer is no longer factual. Our generated example is also counterfactual since the gold answer has also changed over time, but \datasetname{} and TriviaQA examples conflict with each other. The final example showcases ambiguous questions. For this question, we could consider both Niger and Tunisia as potential answers, as they have land borders with Algeria and Libya. However, only the answer `Niger' is attributable because the generated document does not mention Tunisia. The second and third examples illustrate another aspect of our HAR pipeline: HAR can produce conflicting open book QA examples for a given dataset.

\subsection{Evaluation}

We evaluate \datasetname{} in two aspects: attribution and counterfactuality. Following prior work \cite{rashkin_attribution_2023, honovich-etal-2022-true}, we utilize natural language inference (NLI) tasks for evaluation. We use a T5-11B model finetuned with a mixture of NLI, fact verification, and paraphrase detection datasets, MNLI, SNLI, FEVER, PAWS, SciTail, and VitaminC, as proposed in \citet{gao-etal-2023-rarr}. We follow the NLI formulation for open book QA\footnote{premise: \{document\}\textbackslash n\textbackslash n\{question\} hypothesis: \{question\}\textbackslash n\{answer\}} in \citet{chang2023kldivergence, aksitov2023characterizing} and measure attribution and counterfactuality scores as follows:

\textbf{Attribution}: We measure the entailment score when the premise is the generated document and question, and the hypothesis is the question and the generated answer. A high attribution score means that the generated answer is grounded in the generated document.

\textbf{Counterfactuality}: We measure the \textit{contradiction} score when the premise is the generated document and question, and the hypothesis is the question and the \textit{gold answer} in TriviaQA. Since we want \datasetname{} to be counterfactual, we would like there to be no entailment between the generated counterfactual document and the original factual gold answer.

\begin{table}
\centering

\resizebox{0.48\textwidth}{!}{
\begin{tabular}{lll}
\hline
\textbf{} & \textbf{\small{Attribution}} & \textbf{\small{Counterfactuality}}\\
\hline

\datasetname & \textbf{0.77} & \textbf{0.87} \\
\small{- Attribution Filtering} & 0.65 & 0.84 \\
\small{- Factuality Filtering} & 0.68 & 0.65 \\

\hline
\end{tabular}
}
\caption{
Attribution and counterfactuality evaluation of \datasetname{} via NLI. We show that each step of filtering improves the attribution and counterfactuality of our dataset.
}
\label{tab:har-evaluation}
\end{table}

We present the results in Table \ref{tab:har-evaluation} before and after the filtering steps. Without any filtering, the dataset includes many factual examples without text grounding. However, factuality filtering improves counterfactuality by 0.19, and attribution filtering improves overall attribution by 0.12 points.
These qualitative examples and improvements in NLI-based attribution and counterfactuality scores show the importance of the filtering mechanism.

In addition to attribution and counterfactuality scores presented in Table \ref{tab:har-evaluation}, we present qualitative examples labeled as factual by the counterfactual filter in Table \ref{tab:selected_factuals} in the Appendix. These examples illustrate that the surface form heuristics to detect counterfactuals would not be sufficient. We observe a diverse set of factual answer generations that do not have the same surface form as the gold answer, such as hypernyms, ancient names/synonyms, or round numbers. We also illustrate generated examples without grounding according to the attribution filter in Table \ref{tab:selected_attribution} in the Appendix. Although they are all counterfactual examples, they do not have a proper grounding in the text, or there is conflicting information between the question, generated document, and/or generated answer.

\begin{table*}
\centering
\resizebox{0.99\textwidth}{!}{
\begin{tabular}{lrrrrrrrr}
\toprule
\textbf{\makecell[l]{Training\\Dataset}} & \textbf{TriviaQA} &  \textbf{SQuAD} &    \textbf{NQ} &  \textbf{HotpotQA} &  \textbf{BioASQ} &   \textbf{AQA} &  \textbf{AmbigQA} &  \textbf{OOD Avg.} \\
\cmidrule(lr){1-1} \cmidrule(lr){2-2} \cmidrule(lr){3-8} \cmidrule(lr){9-9}
TriviaQA      &      \textbf{85.2} &   79.6 &  66.5 &      69.4 &    63.4 &  42.1 &     \textbf{53.2} &      62.4 \\
\datasetname     &      81.7 &   \textbf{81.7} &  \textbf{71.2} &      \textbf{73.8} &    \textbf{69.5} &  \textbf{44.9} &     \textbf{53.2} &      \textbf{65.7} \\
\bottomrule
\end{tabular}
}
\caption{Token-level F1 scores of T5-3B models finetuned with TriviaQA vs. \datasetname. T5-3B model finetuned with \datasetname{} significantly outperforms T5-3B with TriviaQA by 3.3 points.}
\label{tab:main_results}
\end{table*}

\section{Open book QA Experiments}
\label{sec:experiments}
After describing our experimental setup, we ask 4 important research questions to analyze and measure the effect of counterfactual examples on text grounding by focusing on open book QA experiments.

We finetune T5 models \cite{t5} on either TriviaQA \cite{joshi-etal-2017-triviaqa}, referred to as factual models, or on \datasetname, referred to as counterfactual models, or on the combination of TriviaQA and \datasetname, which is called the combined model. Then, their performance is compared across \textit{out-of-domain datasets}, following the experimental setup in \citet{paranjape-etal-2022-retrieval}. The evaluation includes a comparison across various open book QA datasets, including SQuAD \cite{rajpurkar-etal-2016-squad}, Natural Questions \cite[NQ;][]{kwiatkowski-etal-2019-natural}, HotpotQA \cite{yang-etal-2018-hotpotqa}, BioASQ \cite{Tsatsaronis2015}, AmbigQA \cite{min-etal-2020-ambigqa}, and AQA \cite[adversarially-generated SQuAD questions;][]{bartolo-etal-2020-beat} with the versions of MRQA 2019 shared task \cite{fisch-etal-2019-mrqa}. We mainly present the results with token F1 scores and additionally include the results with the exact match scores in \S\ref{sec:appendix_results} in the Appendix, which exhibit similar trends.

\noindent\textbf{Q1. Does the (hallucination augmented) counterfactual dataset improve text grounding?}

We finetune T5-3B models with human-annotated factual TriviaQA and counterfactual \datasetname{} with Hallucination Augmented Generation (HAR). We compare their performance in various out-of-domain datasets to see the effect of counterfactuals on text grounding.

The results in Table \ref{tab:main_results} show that the counterfactual model achieves a \textbf{3.3} higher token-level F1 score with 4x smaller data than the factual model. It consistently outperforms the factual model on all out-of-domain datasets including multihop, biomedical, and adversarial QA datasets. We see a drop in in-domain performances as expected since the generated dataset conflicts with answers in TriviaQA and our focus is on generalization of text grounding abilities. However, the combined model, which is finetuned on both counterfactual \datasetname{} and factual TriviaQA, achieves both good in-domain performance and out-of-domain performance, as shown in Table \ref{tab:combination_results}.

\begin{table}
\centering
\begin{tabular}{lrrrrrrrr}
\toprule
\textbf{Training Dataset} & \textbf{TriviaQA} &  \textbf{OOD Avg.} \\
\midrule
TriviaQA      &      85.2 &      62.4 \\
\datasetname     &      81.7 &      \textbf{65.7} \\
\midrule
\makecell[l]{TriviaQA\\+\datasetname}  &     \textbf{85.3} &  65.3 \\
\bottomrule
\end{tabular}
\caption{Token-level F1 scores of T5-3B models finetuned with TriviaQA, \datasetname, and their combination. Combining our \datasetname{} dataset with TriviaQA achieves good out-of-domain performance while having a similar performance in in-domain as the model finetuned with TriviaQA.}
\label{tab:combination_results}
\end{table}

\noindent\textbf{Q2. Does the improvement in text grounding via counterfactuals vary with model size?}

We observe a 5.3\% relative improvement between T5-3B models finetuned with \datasetname{} and TriviaQA in Q1. To see how performance improvements vary across different model sizes, we finetune all T5 models, small (60M), base (220M), large (770M), 3B, and 11B.

\begin{figure}[t]
    \centering
    \includegraphics[width=\linewidth]{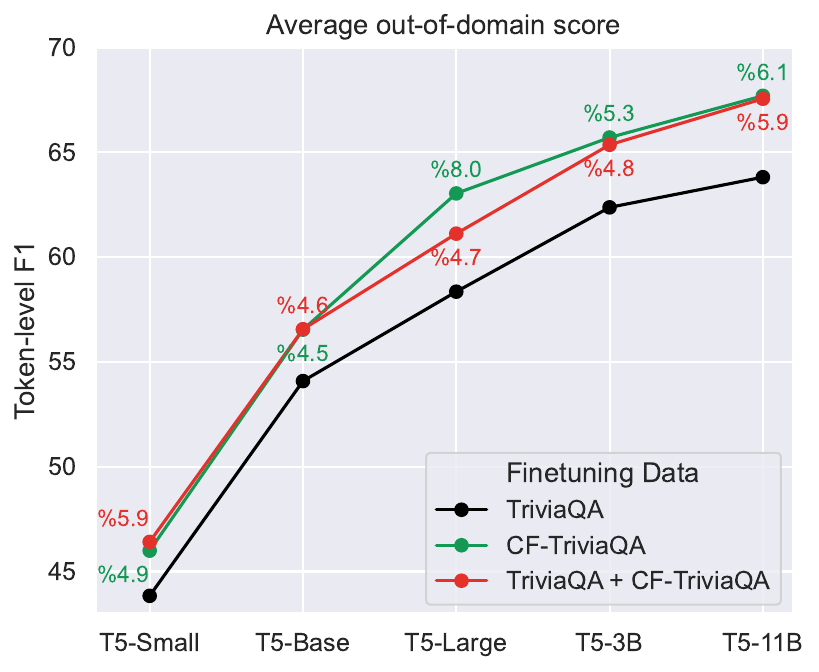}
    \caption{Out-of-domain performance of factual, counterfactual, and combined models with all sizes of T5 models. Models including counterfactual examples consistently outperform factual models across all sizes.
    }
    \label{fig:model_size}
\end{figure}

We show the average F1 score in out-of-domain datasets in Figure \ref{fig:model_size}. We see consistent improvements across all model sizes for both the counterfactual model with \datasetname{} and the combined model with \datasetname{} and TriviaQA over the factual model with TriviaQA. The relative improvement is always between 4.5\% and 8.0\% across all model sizes, suggesting that counterfactuals help to improve language models' text grounding capabilities regardless of the model size.

Furthermore, we can observe that the counterfactual models could achieve much better performance than the larger factual models. In Figure \ref{fig:model_size}, we observe that T5-3B with \datasetname{} outperforms 4x larger model T5-11B with a 4x larger training dataset, with TriviaQA. This verifies the impact of both our HAR approach to generate counterfactual examples and counterfactuals on text grounding.

\noindent\textbf{Q3. What would be the impact on text grounding performance if a factual open book QA dataset, generated through our modified HAR pipeline, was used for finetuning instead of using a counterfactual dataset?}

Our HAR pipeline includes two novel contributions: LLM-generated open book QA dataset, and, more importantly, utilizing hallucination to generate counterfactual examples. To analyze the impact of using an LLM-generated dataset, we generate a factual alternative and compare their performance on out-of-domain open book QA datasets.

We modify the hallucination augmented retrieval pipeline to generate factual open book QA examples which is called \factualdatasetname. After recitation generation, we select factual recitations by comparing the generated answer to the gold answer instead of performing factuality filtering in HAR (i.e., step 2). We still perform attribution filtering (i.e., step 3) with \factualdatasetname{} to get high-quality grounded examples as well. To minimize the impact of other factors, we only include the same questions in \datasetname{} to \factualdatasetname{}. However, there are only around 10K questions common both in \datasetname{} and \factualdatasetname{} due to filtering, therefore we select around 9,000 generated factual questions randomly to keep them the same size.

\begin{table}[t]
    \centering
    \begin{tabular}{lrr}
    \toprule
    \textbf{Training Dataset} &  \textbf{TriviaQA} &   \textbf{OOD Avg.} \\
    \midrule
    TriviaQA     &      \textbf{85.2} &   62.4 \\
    \factualdatasetname &      83.3 &   63.6 \\
    \datasetname     &      81.7 &   \textbf{65.7} \\
    \bottomrule
    \end{tabular}
    \caption{Token-level F1 scores of T5-3B model finetuned with TriviaQA (human-annotated factual), \factualdatasetname{} (LLM-generated factual), \datasetname{} (LLM-generated counterfactual). While \factualdatasetname{} outperforms TriviaQA, showing the strength of LLM generation and our HAR pipeline, \datasetname{} outperforms both factual models, showing the importance of counterfactuals on text grounding. See Table \ref{tab:full_factual_results} in the Appendix for scores in each out-of-domain dataset, separately.}

\label{tab:factual_results}
\end{table}

We compare the performance of T5-3B finetuned with TriviaQA (human-annotated factual), \factualdatasetname{} (LLM-generated factual), and \datasetname{} (LLM-generated counterfactual) on out-of-domain QA datasets. We present the results in Table \ref{tab:factual_results}. We observe that models finetuned with LLM-generated datasets (factual or counterfactual) based on HAR outperform the TriviaQA-based model even with a 4x smaller dataset. This finding is aligned with previous work showing improvements in generating synthetic data using LLMs \cite{wang-etal-2023-self-instruct, koksal2023longform}.

Furthermore, we see much higher improvement from the counterfactual dataset than from the LLM-generated factual dataset when both are compared with TriviaQA. The counterfactual model has a 3.3 higher token-level F1 score, while the model with \factualdatasetname{} has only a 1.2 higher token-level F1 score. This supports our hypothesis that counterfactuals improve text grounding.

\noindent\textbf{Q4. What is the impact of LLM size in Hallucination Augmented Recitations (HAR)? Do smaller models generate the same-quality datasets that lead to similar improvements in text grounding?}

We perform an additional study on the size of LLMs in HAR. We utilize LLM hallucination in the recitation generation of HAR and then filter to get high-quality examples. We replace PaLM 2-Large with a smaller variant, PaLM 2-Small for the recitation generation step while keeping PaLM 2-Large for filtering steps. We generate a new counterfactual dataset with the smaller PaLM model, called \datasetname\textsubscript{PaLM 2-S}, which has the same number of examples as \datasetname\textsubscript{PaLM 2-L}.

\begin{table}
    \centering
    \resizebox{0.48\textwidth}{!}{
    \begin{tabular}{lrr}
    \toprule
    \textbf{Training Dataset} &  \textbf{TriviaQA} &  \textbf{OOD Avg.} \\
    \midrule
    TriviaQA &      \textbf{85.2} &      62.4 \\
    \datasetname\textsubscript{PaLM 2-S}  &      80.8 &      \textbf{66.0} \\
    \datasetname\textsubscript{PaLM 2-L} &      81.7 &     65.7 \\
    \bottomrule
    \end{tabular}
    }
    \caption{Comparison of LLM sizes in the HAR pipeline shows that smaller alternatives of PaLM 2 can achieve similar performance in out-of-domain scores. Therefore, smaller models can be used to utilize hallucination in HAR for better text grounding. See Table \ref{tab:full_small_har_results} in the Appendix for scores in each out-of-domain dataset, separately. }
    \label{tab:small_har_results}
\end{table}

We observe that the hallucination rate in the smaller model is much higher than in the larger model. The initial dataset generated with PaLM 2-S has 31K high-quality counterfactual open book QA examples, which is 50\% more than the data generated by PaLM 2-L. This is consistent with the previous work showing that smaller models tend to hallucinate more than their larger counterparts \cite{elaraby2023halo, rawte2023survey}. For a fair comparison, we again randomly sample the same number of examples (by including the common questions first) as \datasetname\textsubscript{PaLM 2-L}.

We compare their performance on open book QA datasets in Table \ref{tab:small_har_results}. The counterfactual model with \datasetname\textsubscript{PaLM 2-S} achieves even slightly better performance than \datasetname\textsubscript{PaLM 2-L}. This shows that the hallucination generated by smaller language models can be also used for counterfactual data generation via HAR after applying factuality and attribution filtering steps with larger models.

\section{Related Work}
\textbf{Counterfactual Datasets}: Counterfactuals in NLP usually refer to perturbations that make the given text true under different circumstances, while remaining consistent with the possible worlds where the prerequisites hold. Therefore, counterfactuals play a vital role in both the evaluation of language models \cite{qin-etal-2019-counterfactual, wu2023reasoning} and their out-of-domain generalization \cite{bowman-dahl-2021-will} of language models. Prior works on counterfactual generation utilize expensive human annotation \cite{Kaushik2020Learning}, while more recent works focus on automatic generation. Some of these works employ basic heuristics such as negating verbs or swapping noun phrases \cite{dua-etal-2021-learning}, and replacing gendered words in questions \cite{webster-etal-2020}. \citet{paranjape-etal-2022-retrieval} propose a retrieval-based generation system to create counterfactual datasets. There are also some works focusing on perturbing contexts in open book QA with methods such as named entity replacement, thereby changing the answer to create counterfactual examples \cite{longpre-etal-2021-entity, ye-etal-2021-connecting}. However, these methods have difficulty understanding complex structures and may create counterfactual examples that are not consistent (e.g., changing a date of birth without changing the date of death or without changing the occurrence of age in the document). This is reflected in their results, as these approaches have shown only weak and inconsistent improvements in open book QA \cite{paranjape-etal-2022-retrieval, longpre-etal-2021-entity}. In contrast, we propose LLM-based Hallucination Augmented Recitations (HAR) for counterfactual generation. HAR produces high-quality and consistent counterfactual examples, as seen by qualitative examples and out-of-domain performance improvement. To the best of our knowledge, we are the first ones to utilize LLM hallucination to create counterfactual datasets.

\noindent\textbf{Synthetic Data Generation}: Synthetic question answering dataset generation without counterfactuals has shown limited improvement in out-of-domain generalization \cite{bartolo-etal-2021-improving, patrick-etal-2021}. However, recent advancements in large language models (LLMs) have led to growing interest in synthetic data generation with LLMs, such as in more generalized instruction tuning datasets from scratch \cite{wang-etal-2023-self-instruct} or by restructuring existing corpora \cite{koksal2023longform}. Synthetic data generation with LLMs has also been applied to existing datasets for specific tasks to improve model quality, such as natural language inference \cite{liu-etal-2022-wanli} and sentiment analysis \cite{meng-etal-2023-tuning}.

\section{Conclusion}
In this paper, we propose Hallucination Augmented Recitations (HAR) to create a counterfactual open book QA dataset, \datasetname. Since factual open book QA tasks have multi-objective trade-offs (i.e., recalling the answer from the memory of language models 
vs. grounding in the given context), we hypothesize that high-quality counterfactual datasets would further improve attribution. Our results show that models finetuned with \datasetname{} significantly outperform models finetuned with factual TriviaQA, even with a 4x smaller training dataset and a 4x smaller model size. This improvement is consistent across various out-of-domain open book QA tasks, including multi-hop, biomedical, and adversarial questions. For future work, examples in \datasetname{} could help to analyze LLM hallucination as our HAR pipeline enables the generation of high-quality and complex hallucinations. Additionally, the generated counterfactual dataset and the HAR pipeline could be further used to evaluate the text grounding abilities of LLMs \cite{wu2023reasoning} or improve the robustness of natural language inference models, which could later be applied to improve attribution scoring \cite{rashkin_attribution_2023}.

\section*{Acknowledgments}
We would like to thank David Reitter, Yun-hsuan Sung, Hannah Rashkin, and Tania Rojas-Esponda for their valuable feedback.

\bibliography{custom}

\newpage

\appendix

\section{Qualitative Examples from Filter}
\label{sec:appendix_filter_examples}
We present qualitative examples from the factuality filter in Table \ref{tab:selected_factuals} and from the attribution filter in Table \ref{tab:selected_attribution}.

\begin{table*}
\centering
\scriptsize
\begin{tabular}{p{0.15\linewidth} p{0.43\linewidth} p{0.15\linewidth} p{0.15\linewidth}}
\toprule
\textbf{Question:} & \textbf{Generated Document:} & \textbf{Generated Answer} & \textbf{Gold Answer} \\
The element of the Periodic Table LUTETIUM has its name derived from the Latin name for the city in which is was discovered. Which city?
&
Lutetium is a chemical element with symbol Lu and atomic number 71. It is a silvery white metal which resists corrosion in dry, but not moist, air. It is the last element in the lanthanide series, and traditionally counted among the rare earths. Lutetium is usually considered to be the final element in the lanthanide series. However, lutetium has a 4f145d16s2 configuration, which makes it technically the first element of the 5d transition series. Lutetium was named after Lutetia, the Roman name of Paris, where its discovery was announced by Georges Urbain.
&
Lutetia, Paris
&
Paris 
\\
\midrule
\multicolumn{4}{l}{\parbox{0.95\linewidth}{\textbf{Notes:} The generated answer includes the ancient name of Paris, Lutetia, likely to provide more insight into the question about Lutetium.}
}  \\ \midrule
\textbf{Question:} & \textbf{Generated Document:} & \textbf{Generated Answer} & \textbf{Gold Answer} \\ 
What was the population of the world at 31 Oct 2011 according to the United Nations?
&
The world population has been increasing rapidly. Between 1900 and 2000 it increased almost four times. Now it stands at about 6.5 billion. The current projection is that by the year 2025 it will be about 8 billion. United Nations estimates show that the world population at 31 Oct 2011 was 7,001,818,322.
&
7,001,818,322
&
7 billion 
\\
\midrule
\multicolumn{4}{l}{\parbox{0.95\linewidth}{\textbf{Notes:} The generated answer provides more detailed information, but both answers are considered to be the same.}
}  \\ \midrule
\textbf{Question:} & \textbf{Generated Document:} & \textbf{Generated Answer} & \textbf{Gold Answer} \\
The introduction of what vegetable coincided with the coronation of the British monarch in 1902?
&
The year 1902 saw the coronation of King Edward VII and the introduction of the potato crisp to the British public. The crisp is not only the nation’s favourite snack, it is also one of the few British inventions to make it big around the world. In 2004, Britain celebrated the centenary of the crisp, which was the brainchild of a frustrated chef.
&
Potato
&
King Edward potato
\\
\midrule
\multicolumn{4}{l}{\parbox{0.95\linewidth}{\textbf{Notes:} The generated answer is less specific than the gold answer, but both can be considered as factual.}
}  \\ \midrule
\textbf{Question:} & \textbf{Generated Document:} & \textbf{Generated Answer} & \textbf{Gold Answer} \\
Which insect gives off the froth known as cuckoo spit?
&
Spittlebugs are sap-feeding insects, members of the Cercopoidea superfamily. They are commonly called spittlebugs or froghoppers because the nymphs produce a spittle-like protective cover around them. The common name also stems from the fact that many spittlebugs jump well and that they are similar to the true (Ranoidea) froghoppers.
&
Spittlebugs
&
Froghopper
\\
\midrule
\multicolumn{4}{l}{\parbox{0.95\linewidth}{\textbf{Notes:} The parent taxon of the generated answer, Spittlebugs, is the gold answer, Froghopper.}
}  \\

\bottomrule
\end{tabular}
\caption{Examples where the generated answer has a different surface form than the gold answer, but the factuality filter detects the generated answer as factual.}	
\label{tab:selected_factuals}
\end{table*}

\begin{table*}
\centering
\scriptsize
\begin{tabular}{p{0.15\linewidth} p{0.43\linewidth} p{0.15\linewidth} p{0.15\linewidth}}
\toprule
\textbf{Question:} & \textbf{Generated Document:} & \textbf{Generated Answer} & \textbf{Gold Answer} \\
In 2015 which London museum rejected the Thatcher family’s offer to display memorabilia belonging to the former Prime Minister?
&
The museum rejected the Thatcher family’s offer to display memorabilia belonging to the former Prime Minister because it was not “in line with the museum’s collecting policy”. The decision by the museum, which is run by the Imperial War Museums, was made in 2015 but has only now been revealed in a freedom of information (FoI) request made by the Guardian.
&
The Cabinet War Rooms
&
The Victoria and Albert
\\
\midrule
\multicolumn{4}{l}{\parbox{0.95\linewidth}{\textbf{Notes:} The counterfactual document mentions the Imperial War Museums, but the generated answer, `The Cabinet War Rooms', is not grounded in the text, even though it is part of the Imperial War Museums.
}
}  \\ \midrule
\textbf{Question:} & \textbf{Generated Document:} & \textbf{Generated Answer} & \textbf{Gold Answer} \\
Which African country has the shilling as it's currency?
&
The shilling is a unit of currency formerly used in the United Kingdom, Australia, New Zealand and other British Commonwealth countries. The word shilling comes from scilling, an accounting term that dates back to Anglo-Saxon times where it was said to be the value of a cow in Kent or a sheep elsewhere. The shilling was the denomination of the pre-decimal pound sterling, and was worth 12 pence. It was designated by the symbol "s" – not from the first letter of "shilling", but from the Latin solidus. The symbol "s" was used as a shilling abbreviation from the 18th century until decimalisation in 1971.
&
UK
&
Kenya
\\
\midrule
\multicolumn{4}{l}{\parbox{0.95\linewidth}{\textbf{Notes:} The counterfactual document does not mention the UK as a part of Africa, therefore the generated answer is not grounded in the document.
}
}  \\ \midrule

\textbf{Question:} & \textbf{Generated Document:} & \textbf{Generated Answer} & \textbf{Gold Answer} \\
Who was the first British female singer to have a UK No.1 hit in the 1950's?
&
Lonnie Donegan (born Anthony James Donegan; 29 April 1931 – 3 November 2002) was a Scottish musician, known as the "King of Skiffle" and a major figure in the British popular music scene in the 1950s and 1960s. In the late 1950s, he was the first British male singer to have two top ten hits in the US Billboard Hot 100 chart and also the first British artist to have a number one US single. He had two number one hits on the British singles chart.
&
Lonnie Donegan
&
Lita Roza
\\
\midrule
\multicolumn{4}{l}{\parbox{0.95\linewidth}{\textbf{Notes:} The question asks for the first British \textit{female} singer, but the counterfactual document and the generated answer mention the first \textit{male} singer.}}
\\
\bottomrule
\end{tabular}
\caption{Examples where the attribution filter detects the generated answer as not grounded in the generated document. These are all counterfactual examples, since they fed into the attribution filter after the factuality filter.}	
\label{tab:selected_attribution}
\end{table*}

\section{Prompts}
\label{sec:appendix_prompt}

\begin{figure*}
    \centering
    \includegraphics[width=\linewidth]{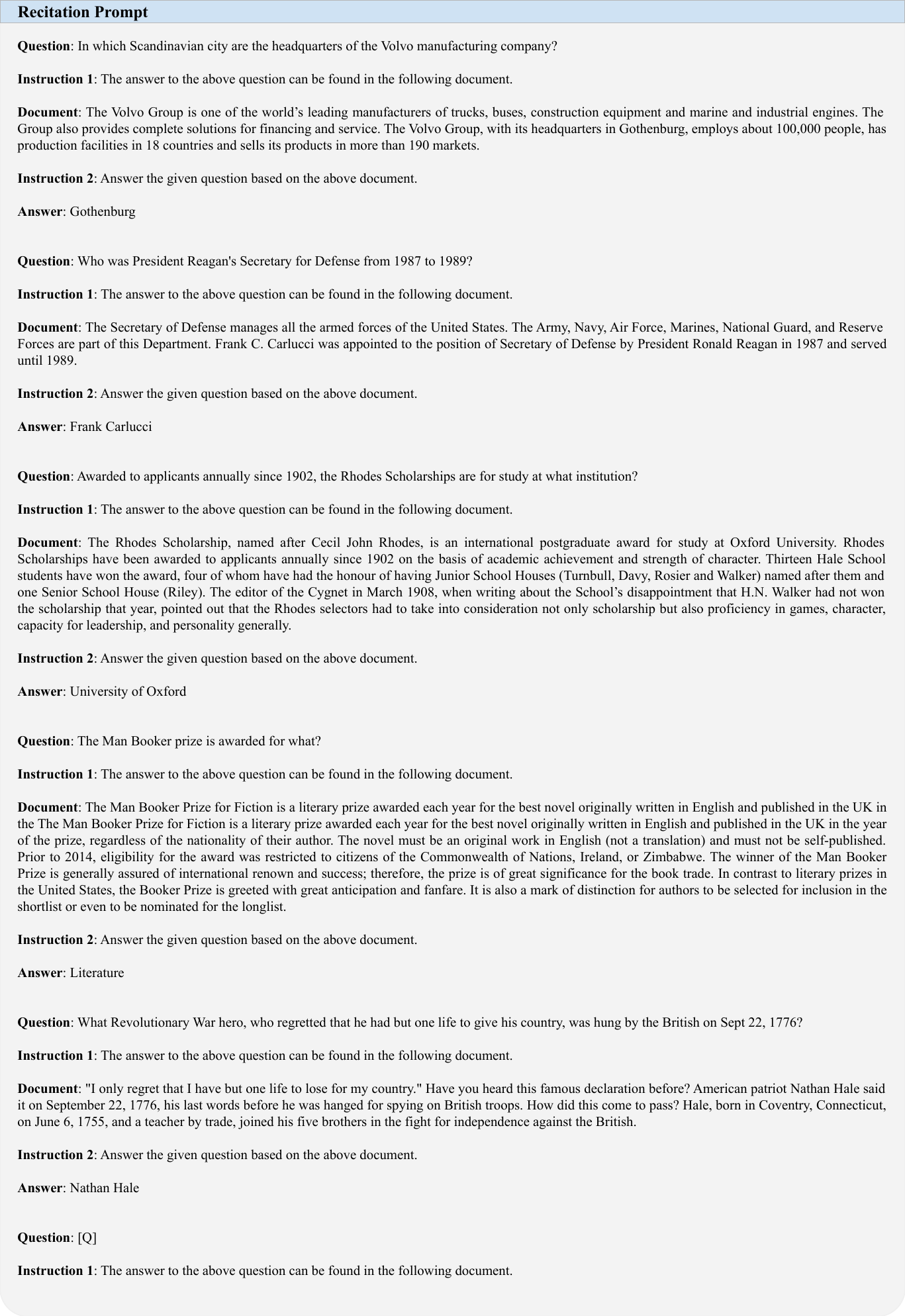}
    \caption{The prompt for the recitation generation step of HAR.}
    \label{fig:recitation_prompt}
\end{figure*}

\begin{figure*}
    \centering
    \includegraphics[width=0.90\linewidth]{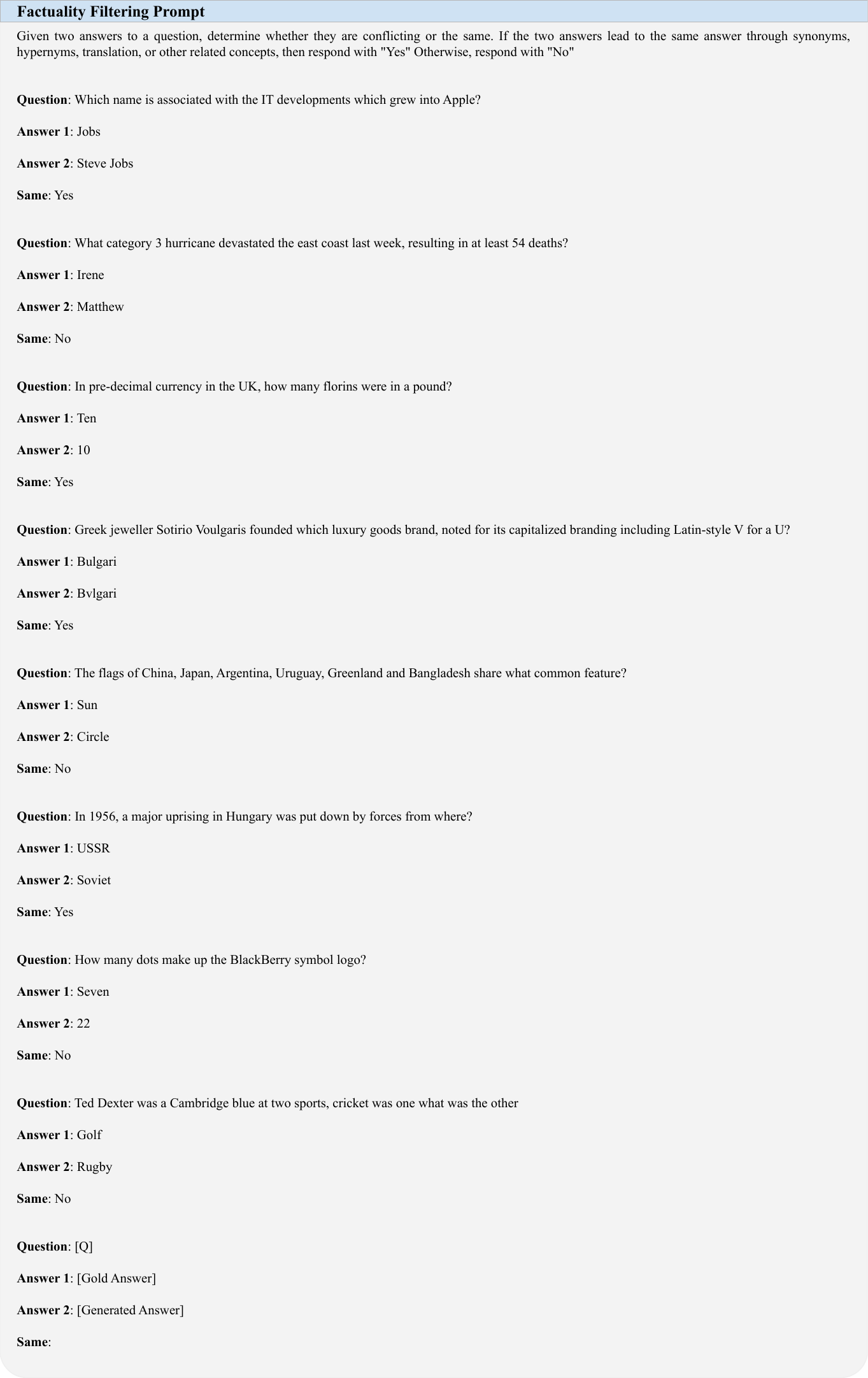}
    \caption{The prompt for the factuality filtering step of HAR.}
    \label{fig:factuality_prompt}
\end{figure*}

\begin{figure*}
    \centering
    \includegraphics[width=\linewidth]{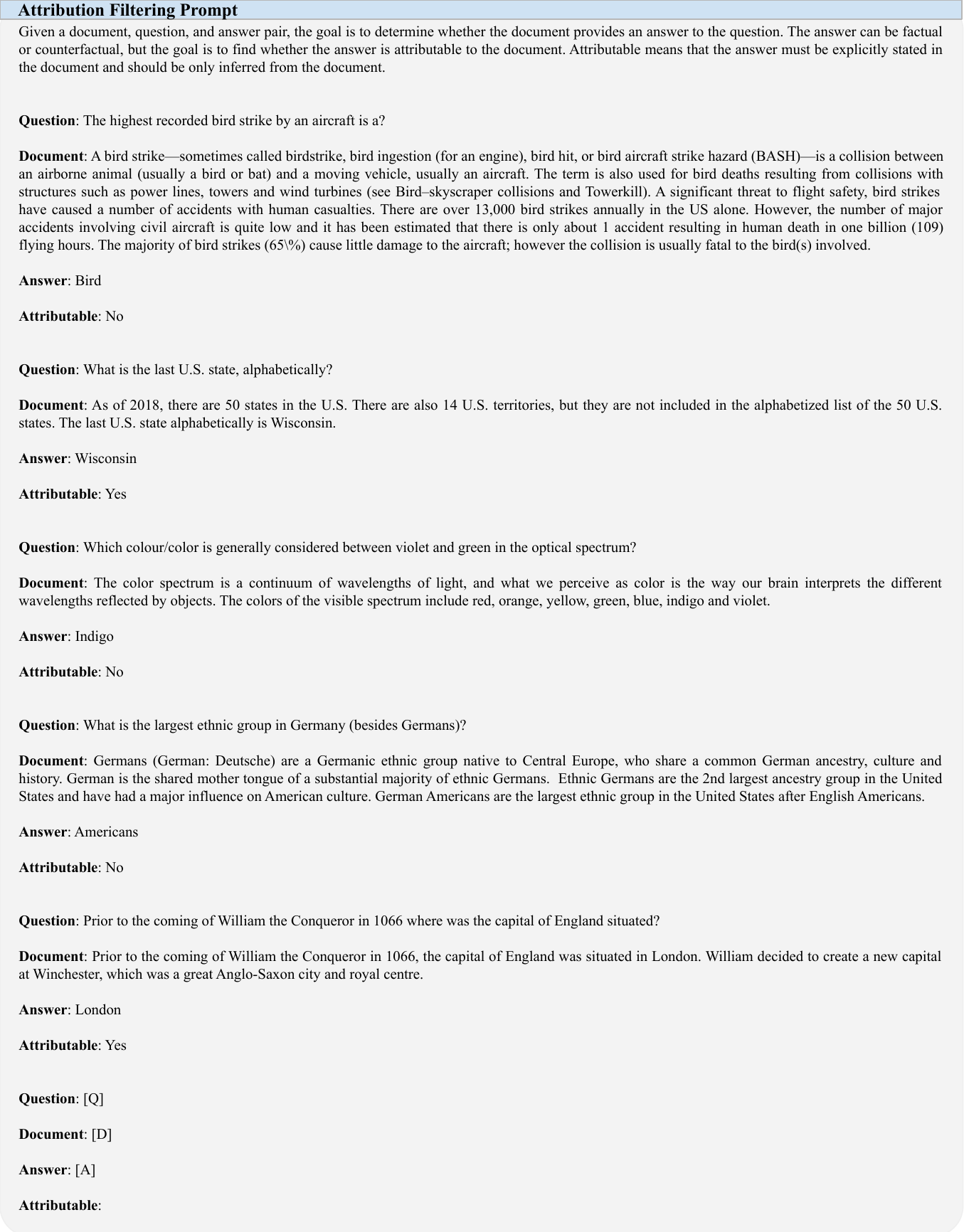}
    \caption{The prompt for the attribution filtering step of HAR. [Q] refers to a question from TriviaQA, [D] and [A] refer to the document and answer pairs generated by LLMs.}
    \label{fig:attribution_prompt}
\end{figure*}

We share the prompts for recitation generation in Figure \ref{fig:recitation_prompt}, for factuality filtering in Figure \ref{fig:factuality_prompt}, and for attribution filtering in Figure \ref{fig:attribution_prompt}. 

\section{Additional Results}
\label{sec:appendix_results}
We share the detailed results with all scores for each out-of-domain dataset for \textbf{Q3} in Table \ref{tab:full_factual_results} and for \textbf{Q4} in Table \ref{tab:full_small_har_results}. 

\begin{figure}[hbt!]
    \centering
    \includegraphics[width=\linewidth]{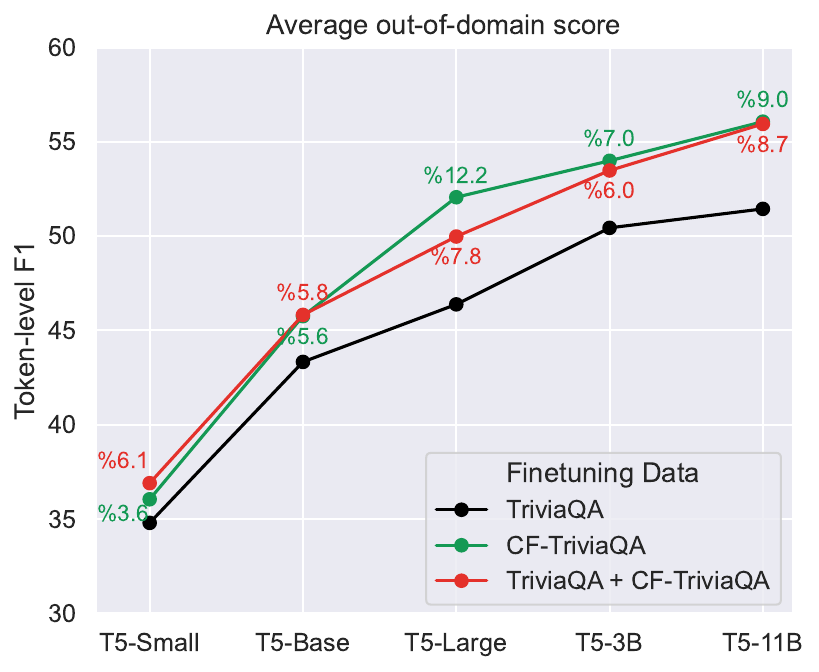}
    \caption{Out-of-domain performance of factual, counterfactual, and combined models with all sizes of T5 models with exact match scores. Models including counterfactual examples consistently outperform factual models across all sizes.
    }
    \label{fig:model_size_em}
\end{figure}

Since we only share token-level F1 scores in \S\ref{sec:experiments}, we present the tables with exact match scores. \textbf{Q1}: Table \ref{tab:combination_results_em} and Table \ref{tab:main_results_em}, \textbf{Q2}: Figure \ref{fig:model_size_em}, \textbf{Q3}: Table \ref{tab:full_factual_results_em}, \textbf{Q4}: \ref{tab:full_small_har_results_em}.

\begin{table}[hbt!]
\centering
\begin{tabular}{lrrrrrrrr}
\toprule
\textbf{Training Dataset} & \textbf{TriviaQA} &  \textbf{OOD Avg.} \\
\midrule
TriviaQA      &      80.9 &      50.4 \\
\datasetname     &      76.6 &      \textbf{54.0} \\
\midrule
\makecell[l]{TriviaQA\\+\datasetname}  &     \textbf{81.0} &  53.5 \\
\bottomrule
\end{tabular}
\caption{Exact match scores of T5-3B models finetuned with TriviaQA, \datasetname, and their combination. Combining our \datasetname{} dataset with TriviaQA achieves good out-of-domain performance while having a similar performance in in-domain as the model finetuned with TriviaQA.}
\label{tab:combination_results_em}
\end{table}

\begin{table*}[hbt!]
\centering
\resizebox{0.99\textwidth}{!}{
\begin{tabular}{lrrrrrrrr}
\toprule
\textbf{\makecell[l]{Training\\Dataset}} & \textbf{TriviaQA} &  \textbf{SQuAD} &    \textbf{NQ} &  \textbf{HotpotQA} &  \textbf{BioASQ} &   \textbf{AQA} &  \textbf{AmbigQA} &  \textbf{OOD Avg.} \\
\cmidrule(lr){1-1} \cmidrule(lr){2-2} \cmidrule(lr){3-8} \cmidrule(lr){9-9}
TriviaQA          &      \textbf{80.9} &   68.6 &  50.5 &      51.9 &    53.3 &  31.6 &     46.8 &      50.4 \\
\datasetname          &      76.6 &   \textbf{70.0} &  \textbf{56.4} &      \textbf{56.7} &    \textbf{60.9} &  \textbf{32.8} &     \textbf{47.1} &      \textbf{54.0} \\
\bottomrule
\end{tabular}
}
\caption{Exact match scores of T5-3B models finetuned with TriviaQA vs. \datasetname. T5-3B model finetuned with \datasetname{} significantly outperforms T5-3B with TriviaQA by 3.6 points.}
\label{tab:main_results_em}
\end{table*}

\begin{table*}
    \centering
    \resizebox{0.99\textwidth}{!}{
    \begin{tabular}{lrrrrrrrr}
    \toprule
    \textbf{Training Dataset} &  \textbf{TriviaQA} &  \textbf{SQuAD} &    \textbf{NQ} &  \textbf{HotpotQA} &  \textbf{BioASQ} &   \textbf{AQA} &  \textbf{AmbigQA} &  \textbf{OOD Avg.} \\
    \cmidrule(lr){1-1} \cmidrule(lr){2-2} \cmidrule(lr){3-8} \cmidrule(lr){9-9}
    TriviaQA     &      \textbf{85.2} &   79.6 &  66.5 &      69.4 &    63.4 &  42.1 &     53.2 &      62.4 \\
    \factualdatasetname &      83.3 &   80.4 &  67.7 &      70.2 &    \textbf{70.2} &  41.8 &     51.3 &      63.6 \\
    \datasetname     &      81.7 &   \textbf{81.7} &  \textbf{71.2} &      \textbf{73.8} &    69.5 &  \textbf{44.9} &     \textbf{53.2} &      \textbf{65.7} \\
    \bottomrule
    \end{tabular}
    }
    \caption{Token-level F1 scores of T5-3B model finetuned with TriviaQA (human-annotated factual), \factualdatasetname{} (LLM-generated factual), \datasetname{} (LLM-generated counterfactual). While models finetuned with both LLM-generated datasets outperform them model with TriviaQA, the counterfactual model significantly outperforms the factual models, demonstrating the importance of counterfactuals in text grounding.}

\label{tab:full_factual_results}
\end{table*}

\begin{table*}
    \centering
    \resizebox{0.99\textwidth}{!}{
    \begin{tabular}{lrrrrrrrr}
    \toprule
    \textbf{Training Dataset} &  \textbf{TriviaQA} &  \textbf{SQuAD} &    \textbf{NQ} &  \textbf{HotpotQA} &  \textbf{BioASQ} &   \textbf{AQA} &  \textbf{AmbigQA} &  \textbf{OOD Avg.} \\
    \cmidrule(lr){1-1} \cmidrule(lr){2-2} \cmidrule(lr){3-8} \cmidrule(lr){9-9}
    TriviaQA     &      \textbf{80.9} &   68.6 &  50.5 &      51.9 &    53.3 &  31.6 &     46.8 &      50.4 \\
    \factualdatasetname &      78.6 &   68.6 &  52.7 &      52.3 &    60.2 &  30.5 &     45.7 &      51.7 \\
    \datasetname     &      76.6 &   \textbf{70.0} &  \textbf{56.4} &      \textbf{56.7} &    \textbf{60.9} &  \textbf{32.8} &     \textbf{47.1} &      \textbf{54.0} \\
    \bottomrule
    \end{tabular}
    }
    \caption{Exact match scores of T5-3B model finetuned with TriviaQA (human-annotated factual), \factualdatasetname{} (LLM-generated factual), \datasetname{} (LLM-generated counterfactual). While models finetuned with both LLM-generated datasets outperform them model with TriviaQA, the counterfactual model significantly outperforms the factual models, demonstrating the importance of counterfactuals in text grounding.}

\label{tab:full_factual_results_em}
\end{table*}

\begin{table*}
    \centering
    \resizebox{0.99\textwidth}{!}{
    \begin{tabular}{lrrrrrrrr}
    \toprule
    \textbf{Training Dataset} &  \textbf{TriviaQA} &  \textbf{SQuAD} &    \textbf{NQ} &  \textbf{HotpotQA} &  \textbf{BioASQ} &   \textbf{AQA} &  \textbf{AmbigQA} &  \textbf{OOD Avg.} \\
    \cmidrule(lr){1-1} \cmidrule(lr){2-2} \cmidrule(lr){3-8} \cmidrule(lr){9-9}
    TriviaQA &      \textbf{85.2} &   79.6 &  66.5 &      69.4 &    63.4 &  42.1 &     53.2 &      62.4 \\
    \datasetname\textsubscript{PaLM 2-S}  &      80.8 &   \textbf{83.3} &  69.7 &      \textbf{73.8} &    \textbf{71.4} &  44.6 &     \textbf{53.2} &      \textbf{66.0} \\
    \datasetname\textsubscript{PaLM 2-L} &      81.7 &   81.7 &  \textbf{71.2} &      \textbf{73.8} &    69.5 &  \textbf{44.9} &     \textbf{53.2} &      65.7 \\
    \bottomrule
    \end{tabular}
    }
    \caption{Comparison of LLM sizes in the HAR pipeline shows that smaller alternatives of PaLM 2 can achieve similar performance in out-of-domain scores with token-level F1 scores. Therefore, smaller models can be used to utilize hallucination in HAR for better text grounding.}

\label{tab:full_small_har_results}
\end{table*}

\begin{table*}
    \centering
    \resizebox{0.99\textwidth}{!}{
    \begin{tabular}{lrrrrrrrr}
    \toprule
    \textbf{Training Dataset} &  \textbf{TriviaQA} &  \textbf{SQuAD} &    \textbf{NQ} &  \textbf{HotpotQA} &  \textbf{BioASQ} &   \textbf{AQA} &  \textbf{AmbigQA} &  \textbf{OOD Avg.} \\
    \cmidrule(lr){1-1} \cmidrule(lr){2-2} \cmidrule(lr){3-8} \cmidrule(lr){9-9}
    TriviaQA &      \textbf{80.9} &   68.6 &  50.5 &      51.9 &    53.3 &  31.6 &     46.8 &      50.4 \\
    \datasetname\textsubscript{PaLM 2-S}  &      75.4 &   \textbf{73.3} &  \textbf{56.6} &      \textbf{57.1} &    \textbf{63.0} &  \textbf{33.7} &     \textbf{47.3} &      \textbf{55.2} \\
    \datasetname\textsubscript{PaLM 2-L} &      76.6 &   70.0 &  56.4 &      56.7 &    60.9 &  32.8 &     47.1 &      54.0 \\
    \bottomrule
    \end{tabular}
    }
    \caption{Comparison of LLM sizes in the HAR pipeline shows that smaller alternatives of PaLM 2 can achieve similar performance in out-of-domain scores with exact match scores. Therefore, smaller models can be used to utilize hallucination in HAR for better text grounding.}

\label{tab:full_small_har_results_em}
\end{table*}

\end{document}